\documentclass[runningheads]{llncs}
\usepackage{color}
\usepackage[T1]{fontenc}

\usepackage{graphicx}
\usepackage{cite}
\usepackage[misc]{ifsym}
\usepackage{subfigure}
\usepackage{hyperref}

\begin{document}
\title{A Task-aware Dual Similarity Network for Fine-grained Few-shot Learning}
\titlerunning{A Task-aware Dual Similarity Network for Fine-grained Few-shot Learning}
% \title{A Task-aware Dual Similarity Network for Few-shot Fine-grained Image Classification\thanks{Supported by organization x.}}
\author{Yan Qi\inst{1} \and
Han Sun\inst{1}{\textsuperscript{(\Letter)}}\and
Ningzhong Liu\inst{1} \and
Huiyu Zhou\inst{2}}

\authorrunning{Y. Qi et al.}

\institute{Nanjing University of Aeronautics and Astronautics, Jiangsu Nanjing, China \and
School of Computing and Mathematical Sciences, University of Leicester, U.K
\email{sunhan@nuaa.edu.cn}\\}

\maketitle  % typeset the header of the contribution
\begin{abstract}
The goal of fine-grained few-shot learning is to recognize sub-categories under the same super-category by learning few labeled samples. Most of the recent approaches adopt a single similarity measure, that is, global or local measure alone. However, for fine-grained images with high intra-class variance and low inter-class variance, exploring global invariant features and discriminative local details is quite essential. In this paper, we propose a Task-aware Dual Similarity Network(TDSNet), which applies global features and local patches to achieve better performance. Specifically, a local feature enhancement module is adopted to activate the features with strong discriminability. Besides, task-aware attention exploits the important patches among the entire task. Finally, both the class prototypes obtained by global features and discriminative local patches are employed for prediction. Extensive experiments on three fine-grained datasets demonstrate that the proposed TDSNet achieves competitive performance by comparing with other state-of-the-art algorithms.
% The abstract should briefly summarize the contents of the paper in

\end{abstract}
\section{Introduction}
% \subsection{A Subsection Sample}
As one of the most important problems in the field of artificial intelligence, fine-grained image classification~\cite{1,2} aims to identify objects of sub-categories under the same super-category. Different from the traditional image classification task~\cite{3,4}, the images of sub-categories are similar to each other, which makes fine-grained recognition still a popular and challenging topic in computer vision.
\par Benefiting from the development of Convolution Neural Networks (CNNs), fine-grained image classification has made significant progress. Most approaches typically rely on supervision from a large number of labeled samples. In contrast, humans can identify new classes with only few labeled examples. Recently, some studies~\cite{5,6} focus on a more challenging setting, which aims to recognize fine-grained images from few samples, and is called fine-grained few-shot learning(FG-FSL).  Learning from fine-grained images with few samples brings two challenges. On the one hand, images in the same category are quite different due to poses, illumination conditions, backgrounds, etc. So how to capture invariant features in limited samples is a particularly critical problem. On the other hand, it is complicated to distinguish subtle visual appearance clues on account of the small differences between categories. Therefore, we consider that the invariant global structure and the discriminative local details of objects are both crucial for fine-grained few-shot classification.
\par

To effectively learn latent patterns from few labeled images, many approaches~\cite{7,8} have been proposed in recent years. These methods can be roughly divided into two branches: the meta-learning methods and the metric learning ones. Metric learning has attracted more and more attention due to its simplicity and effectiveness, and our work will focus on such methods. Traditional approaches such as matching network~\cite{9} and relation network~\cite{10} usually utilize global features for recognition. However, the distribution of these image-level global features cannot be accurately estimated because of the sparseness of the samples. In addition, discriminative clues may not be detected only by relying on global features. CovaMNet~\cite{11} and DN4~\cite{12} introduce the deep local descriptors which are exploited to describe the distribution with each class feature. Furthermore, although these methods learn abundant features, they deal with each support class independently and cannot employ the contextual information of the whole task to generate task-specific features. In conclusion, the importance of different parts changes with different tasks. 
\par In this paper, we propose a Task-aware Dual Similarity Network(TDSNet) for fine-grained few-shot learning, which makes full use of both global invariant features and discriminative local details of images. More specifically, first, a local feature enhancement module is employed to activate discriminative semantic parts by matching the predicted distribution between objects and parts. Second, in the dual similarity module, the proposed TDSNet calculates the class prototypes as global invariant features. Especially, in the local similarity branch, task-aware attention is adopted to select important image patches for the current task. By considering the context of the entire support set as a whole, the key patches in the task are selected and weighted without paying too much attention to the unimportant parts. Finally, both global and local similarities are employed for the final classification. We conduct comprehensive experiments on three popular fine-grained datasets to demonstrate the effectiveness of our proposed method. Especially, our method can also have good performance when there is only one training image.

\section{Related Work}
% Please note that the first paragraph of a section or subsection is
% not indented. The first paragraph that follows a table, figure,
% equation etc. does not need an indent, either.

% Subsequent paragraphs, however, are indented.

\subsubsection{Few shot learning.} Few-shot learning aims at recognizing unseen classes with only few samples. The recently popular literature on few-shot learning can be roughly divided into the following two categories: meta-learning based methods and metric-learning based methods.
\par
Meta-learning based methods attempt to learn a good optimizer to update model parameters. MAML~\cite{14} is dedicated to learning a good parameter initialization so that the model can adapt to the new task after training on few samples. Ravi et al.~\cite{15} propose a meta-learner optimizer based on LSTM to optimize a classifier while also studying an initialization for the learner that contains task-aware knowledge.
\par
Metric-learning based methods aim to measure the similarity by learning an appropriate metric that quantifies the relationship between the query images and support sets. Koch et al.~\cite{16} adopt a siamese convolutional neural network to learn generic image representations, which is performed as a binary classification network. Lifchitz et al.~\cite{17} directly predict classification for each local representation and calculates the loss. DN4~\cite{12} employs k-nearest neighbors to construct an image-to-class search space that utilizes deep local representations. Unlike DN4, which is most relevant to our work, we argue that considering each support class independently may capture features shared among classes that are unimportant for classification. In this paper, task-aware local representations will be detected to explore richer information.
\par
\textbf{Fine-grained image classification.}
Because some early approaches~\cite{18,19} require a lot of bounding boxes or part annotations as supervision that needs a high cost of expert knowledge, more and more researchers are turning their attention to weakly supervised methods~\cite{20,21} that rely only on image-level annotations. Inspired by different convolutional feature channels corresponding to different types of visual modes, MC-Loss~\cite{22} proposes a mutual-channel loss that consists of a discriminality component and a diversity component to get the channels with locally discriminative regions for a specific class. TDSA-Loss~\cite{23} obtains multi-regional and multi-granularity features by constraining mid-level features with the attention generated by high-level features. Different from these methods, we consider that the discriminability of local features obtained only by the attention maps may not be guaranteed. In order to overcome this limitation, the proposed TDSNet activates the local representations with strong discriminability by matching the distribution between the global features and their sub-features, so that the discriminability of global features at fine-grained scales is improved.

\section{Method}

\subsection{Problem Definition}

In this paper, the proposed TDSNet also follows the common setup of other few-shot learning methods. Specifically, few-shot classification is usually formalized as N-way K-shot classification problems. Let $S$ denote a support set that contains $N$ distinct image classes, and each class contains $K$ labeled samples. Given a query set $Q$, the purpose of few-shot learning is classifying each unlabeled sample in $Q$ according to the support set $S$. However, limited samples in $S$ make it difficult to efficiently train a network. Therefore, auxiliary set $A$ is always introduced to learn transferable knowledge to improve classification performance. Note that $S$ and $A$ have their own distinct label spaces without intersections.
\par In order to learn transferable knowledge better, the episode training mechanism~\cite{9} is adopted in the training phase. Specifically, at each iteration, support set $AS$ and query set $AQ$ are randomly selected from auxiliary set $A$ to simulate a new few-shot classification task. In the training process, multiple episodes are constructed to train the model.
\subsection{Overview}
\begin{figure} 
\centering
\includegraphics[width=0.9\textwidth]{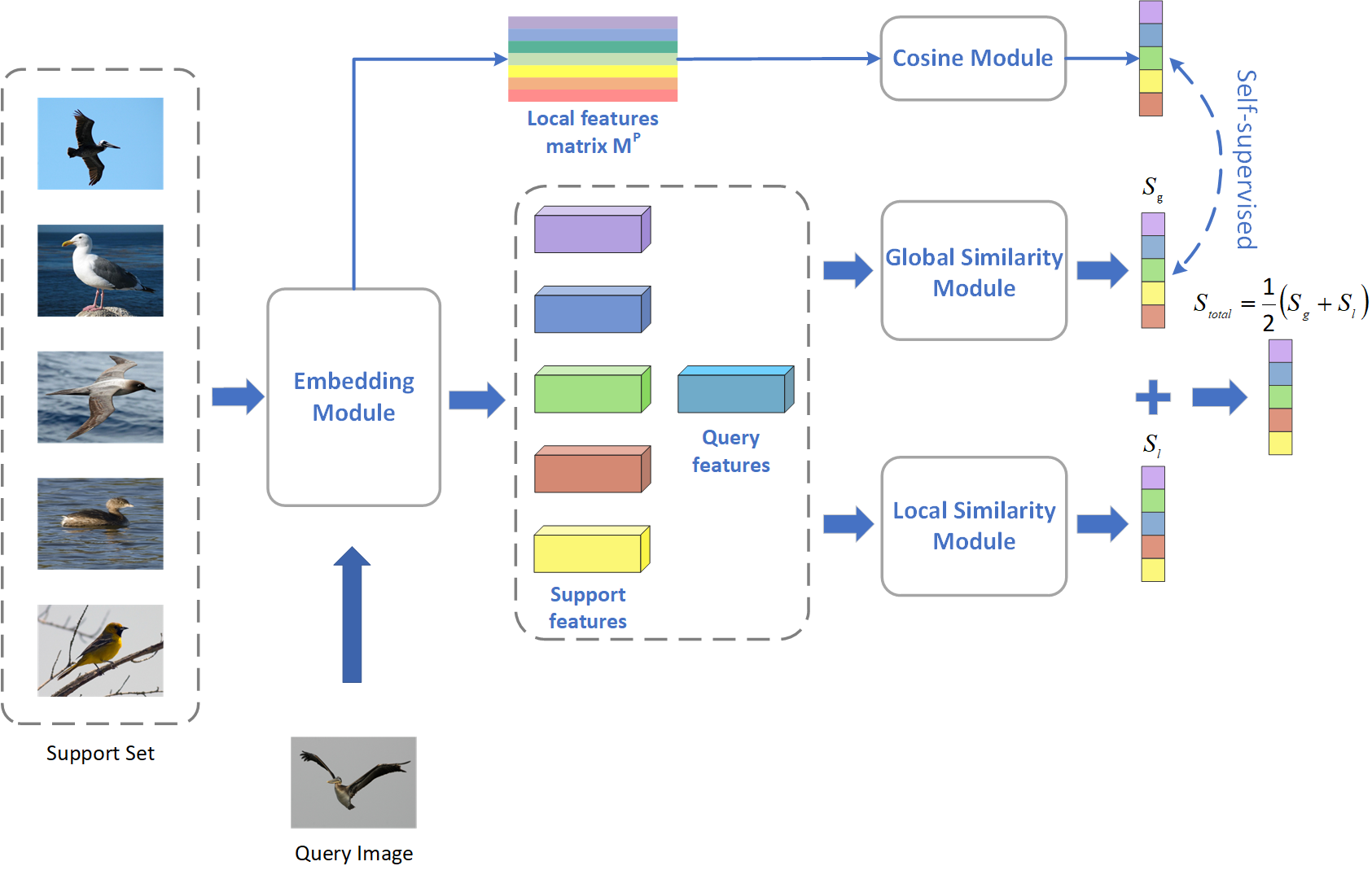}
 %插入图片，[]中设置图片大小，{}中是图片文件名

\caption{Framework of the proposed TDSNet.}

\label{fig2} %用于文内引用的标签
\end{figure}

The overall framework of our method is shown in Fig. \ref{fig2}. First of all, images are fed into the feature extractor which is usually implemented by CNN or ResNet~\cite{24} to get image embeddings. In this stage, a LFE module is designed to explore local details with strong discriminability by the supervision of global features. Next, the features are used as inputs to the metric module. Our metric module adopts dual similarity that is composed of global and local metric branches, which can not only exploit the intra-class invariance of global features but also explore rich clues hidden in local details. Specially, in the local similarity branch, the discriminative patches are reweighted to eliminate noise, i.e., patches shared in the task, and enhance the significant regions. The proposed TDSNet focuses on the relationships among local patches rather than isolated individuals. Finally, the mean value of global and local classification scores is reported as the final result. LFE module is only used during training.
\subsection{Local Feature Enhancement(LFE) Module}
\textbf{Weakly supervised attention generation.}
Parts of the objects are predicted first. In this paper, we explore discriminative regions in a weakly supervised manner. Besides, instead of using a pre-trained convolutional neural network, we employ an attention generation strategy.
\par For the input image $X$, the feature $F\in R^{H\times W\times C}$ is explored through the feature embedding module $f_\varphi$, where $H$, $W$, and $C$ denote the height, width, and the number of channels of the feature respectively. Then, the attention maps $A^a\in R^{H\times W\times m}$ for each image can be determined by
\begin{equation}
A^a = f(F) = \bigcup ^m _{k=1} A^a_k
\end{equation}
$f(\cdot)$ represents a convolution operation and $A^a_k$ is the k-th local attention map. Similar to the bilinear pooling~\cite{25,26}, element-wise multiplication between $A^a$ and $F$ is performed to produce the part feature map $f$, which can be expressed as
\begin{equation}
f_k = g(A^a_k \odot F),k=1,2,\cdot \cdot \cdot,m
\end{equation}
$f_k$ is the k-th local feature, $\odot$  denotes element-wise multiplication, and $g(\cdot)$ is the global average pooling operation. Finally, we stack these part maps to obtain the final part feature matrix. It can be represented as
\begin{equation}
M^P = \left(\begin{array}{c}
    f_1 \\  f_2 \\ \cdot \cdot \cdot \\ f_m 
\end{array}\right)
\end{equation}
\par
\textbf{Local feature enhancement.}
Weakly supervised attention generation is capable of activating some local parts of the objects, however, the discriminability of these local parts may not be guaranteed. Therefore, we propose a feature regularization method to constrain object representation, which extracts knowledge from global features to local features and guides the parts with strong discriminability to be encouraged. An effective way to achieve this effect is to match the prediction distributions between objects and their parts. Let $P_g$ and $P_a$ describe the predicted distributions of the global features and part features $M^P$ respectively. We optimize a KL divergence loss\cite{27} that is applied for measuring the difference between two probability distributions as follows,
\begin{equation}
L_{KL(P_g \parallel P_a)} = -H(P_g) + H(P_g,P_a)
\end{equation}
where $H(P_g) = -\sum P_glogP_g$,  $H(P_g,P_a) = -\sum P_glogP_a$
\par
This regularization loss forces the feature representation learning to focus on the discriminative details from a particular local region, by which we can further filter out unnecessary and misleading information to improve the discriminability of global features at fine-grained scales. Only the one with global features is used for final target prediction.
\subsection{Dual Similarity Module}
\textbf{Global similarity.}
This branch adopts the global feature maps for classification, which employs the cosine distance as a metric function. The feature maps are fed into two convolution blocks($h_{conv}$) to learn generic knowledge with global representations in the images, followed by a cosine similarity $cos(\cdot)$ to measure the similarity. For the query image $X_q$, the global similarity score corresponding to the support class $X_s$ is determined as follows,
\begin{equation}
S_g = cos(h_{conv}(\frac{1}{K}\sum ^K_{s=1}f_\varphi(X_s)),h_{conv}(f_\varphi(X_q)))
\end{equation}
\par
We take the mean value of feature mappings from each support class as the class prototype, which is used to calculate the global structure so that the invariant features within the class can be learned.\par
\begin{figure} 
\centering
\includegraphics[width=0.5\textwidth]{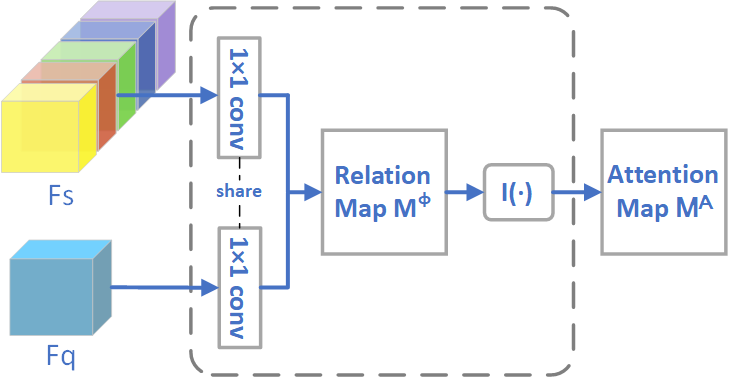}
 %插入图片，[]中设置图片大小，{}中是图片文件名

\caption{The framework of task-aware attention.}

\label{fig3} %用于文内引用的标签
\end{figure}

\textbf{Task-aware local similarity.}
Some recent works, such as DN4 and ConvMNet, have shown that features based on local descriptors are richer than global. Specifically, local descriptors are able to capture local subtle information which is of greater benefit for fine-grained image recognition. For an image feature $F\in R^{H\times W\times C}$, it is regarded as a set of r(r=HW) C-dimensions local feature descriptors, which can be expressed as
\begin{equation}
f_\varphi (X) = [x_1,...,x_r]\in R^{C\times r}
\end{equation}
where $x_i$ denotes the i-th depth local descriptor. These local descriptors correspond to spatial local patches in the raw image. Basically, for each query image $X_q$, we get HW local descriptors to estimate its distribution, denoted as $L^q = f_\varphi(X_q)\in R^{C\times HW}$. Similarly, for support set, all local descriptors of class prototypes will be employed together as $L^s = f_\varphi(X_s)\in R^{C\times NHW}$. Next, we calculate the similarity matrix $M$ between query image and support set by
\begin{equation}
M_{i,j} = cos(L^q_i,L^s_j)
\end{equation}
where $i\in {1,...,HW}$, $j\in {1,...,NHW}$, $cos(\cdot)$ represents cosine similarity. Each row in matrix $M$ represents the similarity of a specific query patch to all patches of the support set. 
\par 
We then construct the task-aware attention map to reweight these parts. As shown in Fig.~\ref{fig3}, we built another relation matrix $M^\phi$ for the next operation, which is obtained by a convolution layer and a cosine similarity layer. We consider that local descriptors shared by multiple classes in the task do not contribute to the classification. For instance, if the query image patch has a high level of similarity to multiple patches in the support set, it has a minuscule contribution to classification. Therefore, we distract attention to make local descriptors that are shared in the task get relatively small attention values. The attention matrix $M^A$ is defined as
\begin{equation}
M_{i,j}^A = \frac{I(M^\phi_{i,j})}{\sum_j I(M^\phi_{i,j})}
\end{equation}
\begin{equation}
I(x) =  \left\{ 
        \begin{array}{ll}
         x,& if \  x > \beta \\
         0,& otherwise.
         \end{array} \right .
\end{equation}
$\beta$ is the threshold, which is set as the minimum of the top-k elements obtained by k-NN~\cite{13} from the relationship matrix $M^\phi$ to eliminate the noises. Since $I(\cdot)$ is indifferentiable, we approximate it by a variant function of sigmoid with a hyperparameter t as
\begin{equation}
I^*(x) = x/(1+exp^{-t(x-\beta)})
\end{equation}
Theoretically, when t is large enough, it can be approximated as $I(\cdot)$. We perform element-wise multiplication between the weight matrix $M^A$ and the relation matrix $M$. Finally, the local similarity score for n-th class between the query image $X_q$ and the support class $X_s$ can be calculated by applying the attention map to the similarity matrix $M$ as follows:
\begin{equation}
S_l = \frac{1}{HW}\sum^{HW}_{i=1}\sum^{HW}_{j=1}(M^A \odot M)_{i,j}
\end{equation}
\par
The total classification score is formulated as follows which is used to make a final prediction:
\begin{equation}
S_{total} = \frac{1}{2}(S_g + S_l)
\end{equation}
\par
In particular, our local similarity branch only introduces a small number of parameters and the overfitting problem in the few-shot learning can also be alleviated to some extent.
\subsection{Loss Function}
In the training phase, the purpose is to learn a task agnostic network for classification. We can obtain $y^g_q$ and $y^l_q$ as two predicted results by global and local branches respectively. Then, predicted values are compared with the ground-truth label $y_q$ to calculate two classification losses. 
\begin{equation}
L^g_q = \sum ^N_j(y^g_{q,j}-y_{q,j})^2, q=1,...,|Q|
\end{equation}
\begin{equation}
L^l_q = \sum ^N_j(y^l_{q,j}-y_{q,j})^2, q=1,...,|Q|
\end{equation}
\par
In the end, the whole loss function can be written as: 
\begin{equation}
L_{total} = L^g_q + L^l_q + \lambda L_{KL}
\end{equation}
Where $\lambda $ is a trade-off parameter used to control the relative importance of the loss $L_{KL}$. Empirically, we set $\lambda  = 0.4$.

\section{Experiment}
\subsection{Datasets and Experimental Setting}
We evaluate our method on three widely used fine-grained datasets, namely CUB-200-2011~\cite{28}, Stanford Dogs~\cite{29}, and Stanford Cars~\cite{30}. We conduct experiments under the 5-way 1-shot and 5-way 5-shot settings. All images are resized to 84 × 84 before being fed into the feature extraction module. In the training process, episode training mechanism is used to train our model. For the three datasets, models are trained for 600 and 400 epochs corresponding to the 5-way 1-shot and 5-way 5-shot tasks, respectively. We use Adam optimizer to train the network with the initial learning rate of 0.001, decaying by half every 100,000 episodes. In the testing phase, the top-1 accuracy with 95\% confidence interval will be reported by random sampling of 600 episodes from the test set.

\begin{table}
\caption{Comparison with typical FSL and FG-FSL methods on three fine-grained datasets. The best and the second best
results are highlighted in red and green respectively.}\label{tab1}

\begin{tabular}{l|c c c c c c}

\hline
{Dataset} & \multicolumn{2}{c}{CUB Birds} 
& \multicolumn{2}{c}{Stanford Dogs} & \multicolumn{2}{c}{Stanford Cars}\\
\hline

{Setting} &\makebox[0.15\textwidth][c]{1-shot}&\makebox[0.13\textwidth][c]{5-shot}&
\makebox[0.15\textwidth][c]{1-shot} &\makebox[0.13\textwidth][c]{5-shot} &\makebox[0.15\textwidth][c]{1-shot}&\makebox[0.13\textwidth][c]{5-shot}\\
\hline
Matching Net~\cite{9}& 45.30±1.03 & 59.50±1.01 & 35.80±0.99 & 47.50±1.03 & 34.80±0.98 & 47.50±1.03\\
Prototype Net~\cite{33}& 37.36±1.00
 & 45.28±1.03 & 37.59±1.00 & 48.19±1.03 & 40.90±1.01 & 52.93±1.03\\
Relation Net~\cite{10}& 59.58±0.94 & 77.62±0.67 & 43.05±0.86 & 63.42±0.76 & 45.48±0.88 & 60.26±0.85\\
MAML~\cite{14}& 54.92±0.95 & 73.18±0.67 & 44.64±0.89 & 60.20±0.80 & 46.71±0.89 & 60.73±0.85\\
PCM~\cite{32}& 42.10±1.96 & 62.48±1.21 & 28.78±2.33 & 46.92±2.00 & 29.63±2.38 & 52.28±1.46\\
CovaMNet~\cite{11}& 52.42±0.76 & 63.76±0.64 & 49.10±0.76 & 63.04±0.65 & 56.65±0.86 & 71.33±0.63\\
DN4~\cite{12}& 46.84±0.81 & 74.92±0.64 & 45.41±0.76 & 63.51±0.62 & \textcolor{green}{59.84±0.80} & \textcolor{red}{88.65±0.44}\\
BSNet~\cite{31}& 65.89±1.00 & 78.48±0.65 & \textcolor{green}{51.68±0.95} & 67.93±0.75 & 54.39±0.92 & 73.37±0.77\\
FOT~\cite{5}& \textcolor{green}{67.46±0.68} & \textcolor{red}{83.19±0.43} & 49.32±0.74 & \textcolor{green}{68.18±0.69} & 54.55±0.73 & 73.69±0.65\\
\hline
ours&	\textcolor{red}{69.34±0.89}	&\textcolor{green}{80.34±0.59}&	\textcolor{red}{54.48±0.87}&	\textcolor{red}{69.45±0.69}&	\textcolor{red}{62.14±0.91}&	 \textcolor{green}{75.64±0.72}\\

\hline

\end{tabular}
\end{table}
\subsection{Comparison with State-of-the-art Methods}
To evaluate the validity of the proposed TDSNet, we conduct extensive experiments on three classic fine-grained datasets with 5-way 1-shot and 5-way 5-shot task settings and compare them with some SOTA methods.
\par
As is demonstrated in Table.~\ref{tab1}, the results compared with four classical few-shot methods and five fine-grained few-shot methods illustrate that our method achieves good performance on all three datasets. Specifically, the proposed TDSNet performs better on the more challenging 1-shot task and shows high stability. The reason for this progress is that our approach focuses on the discriminative parts and gives them higher weights. Additionally, the changes in visual appearance may not affect our TDSNet because we also pay attention to invariant global structure.

\subsection{Ablation Study}

\begin{table}
\centering
\caption{Ablation study on the proposed components on CUB. LFE: local feature enhancement module, LS: local similarity module, att: task-aware attention.}\label{tab2}

\begin{tabular}{l|c | c}
\hline
 \makebox[0.18\textwidth][l]{Method} & \makebox[0.18\textwidth][c]{1-shot} & \makebox[0.18\textwidth][c]{5-shot} \\\hline
(a)Baseline &	63.33±1.01 &	77.64±0.67\\
(b)Baseline+LFE &	65.20±0.99 &	78.77±0.67\\
(c)Baseline+LFE+LS w/o att &	67.02±0.96 &	79.59±0.64\\
(d)Baseline+LFE+LS w/ att &	69.34±0.89	 & 80.34±0.59
\\
\hline
\end{tabular}
\end{table}

\textbf{Effectiveness of local feature enhancement module.}
We use the feature embedding module and the global similarity metric module of this paper as the baseline. Table.~\ref{tab2} shows that the addition of the local enhancement module makes the accuracy significantly improved by 1.87\% and 1.13\% respectively, which is mainly due to the activation of the features with strong discriminability. In this way, we can further filter out misleading information so as to improve the discrimination of features.\par

\textbf{Effectiveness of dual-similarity.} Compared with the results that only employ the global similarity measurement module, dual similarity demonstrates its superiority. It proves that only global features are hard to detect some detailed information that is suitable for fine-grained features, while only local features are sensitive to some intra-class changes. \par
\textbf{Effectiveness of task-aware attention.} We verify the effectiveness of task-aware attention on CUB, with 2.32\% and 0.75\% improvements respectively. Task-aware attention makes the network pay more attention to the features that are most relevant to the current task and reweight the key parts so that the features shared between classes will obtain less attention.

\subsection{Visualization} 
\begin{figure}
\centering
\includegraphics[width=0.65\textwidth,height=0.4\textwidth]{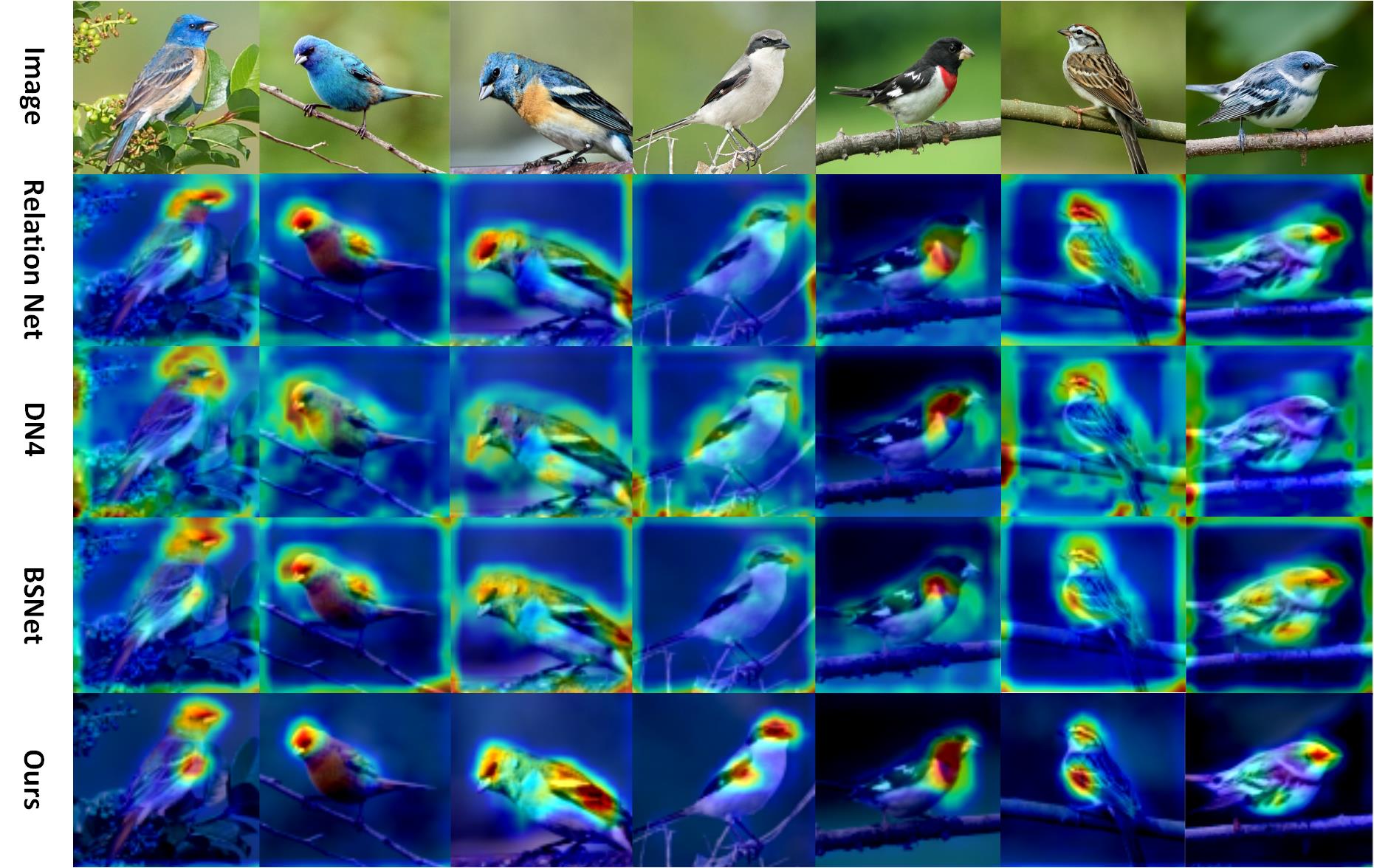}
\caption{ Visualization of the features under Relation Network, DN4, BSNet and the proposed TDSNet on CUB. The redder the region, the more discriminative it is.} %最终文档中希望显示的图片标题
\label{fig4} %用于文内引用的标签
\end{figure}

As can be seen from Fig.~\ref{fig4}, our TDSNet has less activation in the background and is more concentrated on the discriminative regions of objects, which demonstrates that the features can be semantically enhanced by our approach. It highlights the local details on raw images that represent different semantic patches with strong discriminability, such as the head and wings of a bird.

\subsection{Number of Trainable Parameters}
As Table.~\ref{tab3} shows, compared to the BSNet, which is also a bi-similarity method on the global feature, we have only half the number of parameters of this approach. This illustrates that although we adopt additional architectures to improve the performance, the proposed TDSNet only introduces a small number of the trainable parameters.
\begin{table}
\centering
\caption{Comparsion of the number of trainable parameters along on CUB.}\label{tab3}
\begin{tabular}{l| c | c}

\hline
 \makebox[0.18\textwidth][l]{Model} & \makebox[0.18\textwidth][c]{Params} & \makebox[0.18\textwidth][c]{5-way 5-shot} \\\hline
Prototype network &	0.113M	&45.28 \\
Relation network&	0.229M&	77.62\\
DN4 &	0.113M	& 74.92\\
BSNet(R\&C) &	0.226M &	78.84\\
\hline
TDSNet(ours) &	0.114M &	80.34\\

\hline
\end{tabular}
\end{table}
\section{Conclusion}
In this paper, we propose a Task-aware Dual Similarity Network (TDSNet) for FG-FSL, which consists of two designed components, a local feature enhancement module and a measurement module that combines global and task-aware local similarity. Specifically, the former is designed to fully explore the discriminative details suitable for fine-grained classification, while the latter explores similarity by taking multiple perspectives, both global and local. Extensive experiments demonstrate that our proposed TDSNet achieves competitive results. In the future, we intend to reinforce the features of the foreground object and eliminate the negative effect of complicated backgrounds.
\par
\par

\bibliographystyle{splncs04}
\bibliography{ref}

\end{document}